\documentclass{article}
\usepackage[numbers,square,comma,sort]{natbib}
\usepackage[left=2cm,right=2cm, top=3cm]{geometry}
\usepackage[parfill]{parskip}
\usepackage[utf8]{inputenc}
\usepackage[T1]{fontenc}
\usepackage{amssymb}
\usepackage{amsmath}
\usepackage{amsbsy}
\usepackage{amsfonts}
\usepackage[nice]{nicefrac}
\usepackage{breqn}
\usepackage{mathtools}
\usepackage{graphicx}
\usepackage{subcaption}
\usepackage{svg}
\usepackage{tikz}
\usepackage{tikz-cd}
\usepackage{pgfplots}
\usepackage{float}
\usepackage[hidelinks]{hyperref}
\usepackage{booktabs}
\usepackage{microtype}
\usepackage{multirow}
\usepackage{booktabs}
\usepackage{adjustbox}
\usepackage{soul}
\usepackage{wrapfig}
\usepackage{xcolor}
\definecolor{princetonorange}{HTML}{FF8F00}
\hypersetup{pdfborder={0 0 0}}  % no boxes
\makeatletter
\AtBeginDocument{%
  \def\Hy@colorlink#1{\begingroup\color{princetonorange}}%
  \def\Hy@endcolorlink{\endgroup}%
}
\makeatother
\allowdisplaybreaks
\pgfplotsset{compat=1.16}
\newcommand\blfootnote[1]{%
  \begingroup
  \renewcommand\thefootnote{}\footnotetext{#1}%
  \addtocounter{footnote}{0}%
  \endgroup
}
\begin{document}
\title{Demonstration Calibrated Port-Hamiltonian Retuning\\ for Manipulation Policies}
%%%%%%%%%%%%%%%%%%%%%%%%%%%%%%%%%%%%%%%%%%%%%%%%%%%%%%%%
\author{Yulong Yang $^{1,\dagger}$ \quad Fan Wu $^{2}$ \quad Christine Allen-Blanchette $^{1}$ \quad Amit Chakraborty $^{3,\dagger}$\\
$^{1}$Princeton University \quad $^{2}$Arizona State University \quad $^{3}$Siemens Foundation Technologies}
\date{}
\begin{NoHyper}
\blfootnote{Preprint, under review. This work was primarily done during Yulong Yang and Fan Wu's internship at Siemens Foundation Technologies, Princeton, NJ. $^{\dagger}$Correspondence to \texttt{amit.chakraborty@siemens.com} and \texttt{yulong.yang@princeton.edu}.}
\end{NoHyper}
%%%%%%%%%%%%%%%%%%%%%%%%%%%%%%%%%%%%%%%%%%%%%%%%%%%%%%%%
\maketitle
%%%%%%%%%%%%%%%%%%%%%%%%%%%%%%%%%%%%%%%%%%%%%%%%%%%%%%%%
\begin{abstract}
\noindent Diffusion and VLA policies for manipulation are often deployed through downstream impedance controllers. The stiffness and damping gains of these controllers affect task success, yet are commonly inherited from data collection rather than selected for the deployed policy. Although empirical gain sweeps can improve performance, they require repeated evaluation rollouts. We introduce PHRetune, an offline method that derives controller gains for a frozen policy without evaluation rollouts or gain search. Our approach learns a port-Hamiltonian model from demonstrations to estimate the effort and energy associated with the policy’s predicted actions. The policy is applied to recorded demonstration observations, and its predictions are assessed against demonstration-derived effort and energy budgets. From this comparison, we derive a single gain scale in closed form, adjusting the downstream controller while preserving the policy and its action representation. The gains are fixed before evaluation, without requiring a prior manipulator model, task rewards, policy retraining, or additional runtime computation. Across LIBERO suites, PHRetune improves Diffusion Policy success by up to 9.4 percentage points, with the derived gains achieving the highest observed success rates in empirical gain sweeps. On all four real-world manipulation tasks, PHRetuned Diffusion Policy outperforms the nominal policy, alternative gain-tuning methods, and a policy-retraining baseline. The same procedure improves success with SmolVLA and OpenVLA-OFT on every task, while reducing acceleration and jerk for both VLA backbones.
\end{abstract}
%%%%%%%%%%%%%%%%%%%%%%%%%%%%%%%%%%%%%%%%%%%%%%%%%%%%%%%%
\section{Introduction}\label{sec:introduction}
The behavior of a learned manipulation policy depends not only on its predicted actions, but also on the controller that executes them.
Diffusion and vision-language-action (VLA) policies often command end-effector targets through a downstream impedance or operational-space controller~\citep{chi2025diffusion,shukor2025smolvla,kim2025fine}.
The controller’s stiffness and damping therefore shape how policy predictions are realized as motion and interaction forces~\citep{hogan1984impedance,khatib1987unified}.
Yet these gains are commonly inherited from teleoperated data collection or simulator defaults rather than selected for the deployed policy~\citep{zhu2020robosuite,mandlekar2021matters,liu2023libero}.
Recent studies show that this choice materially affects policy learning and execution, and that more compliant settings can improve task success~\citep{bronars2026tune,xu2026mind}.

Prior work addresses this issue either by incorporating compliance into policy learning or by tuning the downstream controller.
Variable-impedance actions, explicit compliance predictions, and learning from intent–execution mismatch modify the action representation or training procedure rather than retuning the controller of an unchanged policy~\citep{martin2019variable,hou2025adaptive,xu2026mind}.
Controller-side approaches instead optimize tracking through a dynamics model, prescribe stiffness from demonstration variability, or search directly over evaluation rollouts~\citep{cheng2024difftune,calinon2010learning,bronars2026tune}.
These approaches respectively rely on a model-based tracking objective, omit the deployed policy’s predictions, or require execution feedback to select gains.
Our goal is to derive a fixed gain tailored to a frozen policy using its task demonstrations, without requiring a prior manipulator model, policy retraining, or gain-selection rollouts.

We introduce PHRetune (pronounced ``free-tune"), an offline method that derives controller gains for a frozen policy from its existing task demonstrations, without additional robot interaction for gain selection.
A port-Hamiltonian formulation relates action attenuation and controller softening through their effects on the energy supplied by policy-generated setpoints~\citep{hogan1984impedance,van2004port}.
Under a linear impedance model, this relationship provides a closed-form map from admissible action scalings to controller gains. 
We determine this action scaling using a surrogate fitted to demonstration transitions, assessing the policy’s predictions on recorded observations against demonstration-derived effort and energy budgets. 
Gain selection therefore remains blind to evaluation outcomes, requiring neither gain search nor additional robot or simulator rollouts. 
The resulting controller gains are fixed before evaluation, leaving the policy, action representation, and inference procedure unchanged and introducing no additional runtime overhead.

Across three LIBERO suites, PHRetuned Diffusion Policy outperforms the nominal policy, alternative gain-tuning methods, and a policy-retraining baseline. 
The gains derived before evaluation also achieve the highest observed success rates in empirical gain sweeps. 
These improvements extend to four real-world manipulation tasks, where PHRetuned Diffusion Policy again outperforms the nominal policy and all evaluated baselines on every task.
Applying the same procedure to SmolVLA and OpenVLA-OFT improves success on every task, with motion evaluations showing reduced acceleration and jerk for both VLA backbones.
%%%%%%%%%%%%%%%%%%%%%%%%%%%%%%%%%%%%%%%%%%%%%%%%%%%%%%%%
%%%%%%%%%%%%%%%%%%%%%%%%%%%%%%%%%%%%%%%%%%%%%%%%%%%%%%%%
\section{Related Work}\label{sec:relatedwork}
%%%%%%%%%%%%%%%%%%%%%%%%%%%%%%%%%%%%%%%%%%%%%%%%%%%%%%%%
\paragraph{Controller Gains in Policy Learning.}
Visuomotor manipulation policies, including diffusion policies and VLA models, commonly predict joint- or end-effector targets that are executed by lower-level controllers~\cite{zhao2023learning,chi2025diffusion,shukor2025smolvla,kim2024openvla}. 
Their closed-loop behavior therefore depends on controller gains, although benchmark environments typically treat these gains as fixed parameters rather than quantities selected for the deployed policy~\citep{james2020rlbench,zhu2020robosuite,mandlekar2021matters,liu2023libero,fei2025libero,nasiriany2024robocasa}.

One line of work incorporates compliance into policy learning. VICES jointly commands Cartesian motion and impedance~\citep{martin2019variable}, while related methods learn variable-impedance profiles from task experience or demonstrations~\citep{buchli2011learning,kronander2013learning}.
Adaptive Compliance Policy predicts compliance alongside diffusion-generated motion~\citep{hou2025adaptive}, and Mind the Gap learns implicit impedance from intent–execution mismatch~\citep{xu2026mind}.
These approaches modify the action representation or training procedure to learn compliant behavior, rather than retuning the downstream controller of an unchanged policy.

A complementary line tunes the downstream controller.
DiffTune differentiates a tracking objective through a dynamics model~\citep{cheng2024difftune}, while demonstration-based rules map motion variability to stiffness~\citep{calinon2010learning}.
Tune-to-Learn empirically evaluates how controller gains affect policy learning and execution, demonstrating that gain selection can substantially alter task performance~\citep{bronars2026tune}.
These approaches select gains through tracking objectives, demonstration statistics, or evaluation rollouts.
PHRetune instead uses a demonstration-calibrated energy model to assess the frozen policy’s predictions on recorded observations against demonstration-derived effort and energy budgets.
This assessment determines a fixed gain scale while preserving the policy, its action representation, and the controller structure.
Gain selection therefore requires no policy retraining nor robot or simulator rollouts.
%%%%%%%%%%%%%%%%%%%%%%%%%%%%%%%%%%%%%%%%%%%%%%%%%%%%%%%%
\paragraph{Impedance, Passivity, and Port-Hamiltonian Foundations.}
Impedance control specifies the dynamic relationship between motion and interaction force around a virtual equilibrium~\citep{hogan1984impedance}, while operational-space control formulates motion and force control directly in task coordinates~\citep{khatib1987unified}.
When a learned policy commands this equilibrium, its setpoint updates can supply energy to the controlled system rather than serving solely as kinematic targets. Port-Hamiltonian modeling describes this interaction through stored energy, dissipation, and power-conjugate ports~\citep{van2004port}.
Under the linear impedance model considered here, spring effort and setpoint velocity form a power-conjugate pair.
This representation provides a common basis for relating action attenuation and controller softening through their effects on the energy associated with setpoints.

These foundations also underpin energy-shaping and passivity-based control. 
Interconnection and damping assignment passivity-based control (IDA-PBC) shapes the closed-loop storage and dissipation structure through feedback synthesis~\citep{ortega2002interconnection}, whereas energy-tank methods regulate energy exchange during online variations in stiffness or interaction~\citep{ferraguti2013tank,schindlbeck2015unified}. 
PHRetune addresses a different design problem.
It neither synthesizes a new feedback law nor enforces a runtime energy budget, but uses a demonstration-calibrated energetic model to derive fixed gains for an existing impedance controller before deployment.
The port-Hamiltonian formulation thus supports an offline retuning rule tailored to a frozen policy, with gain selection remaining blind to evaluation outcomes.
%%%%%%%%%%%%%%%%%%%%%%%%%%%%%%%%%%%%%%%%%%%%%%%%%%%%%%%%
%%%%%%%%%%%%%%%%%%%%%%%%%%%%%%%%%%%%%%%%%%%%%%%%%%%%%%%%
\section{Method}\label{sec:method}
%%%%%%%%%%%%%%%%%%%%%%%%%%%%%%%%%%%%%%%%%%%%%%%%%%%%%%%%%%%
Given a frozen diffusion or vision-language-action policy, its task demonstrations, and a downstream operational-space impedance controller, our goal is to derive fixed controller gains without policy retraining or evaluation rollouts.
PHRetune uses two complementary port-Hamiltonian models to connect offline assessment of the policy’s actions to controller retuning.
An action-port surrogate fitted to demonstration transitions provides model-relative effort and energy estimates.
We apply the frozen policy to recorded demonstration observations and assess its predicted action chunks against demonstration-derived budgets to determine an admissible action scale.
A linear impedance setpoint-port model then relates action attenuation to controller softening through energy matching, providing a closed-form map from this scale to controller gains.
The resulting gains are fixed before evaluation and introduce no additional runtime computation.
%%%%%%%%%%%%%%%%%%%%%%%%%%%%%%%%%%%%%%%%%%%%%%%%%%%%%%%%%%%
\subsection{Port-Hamiltonian Surrogate}
Conservative Hamiltonian models provide an energy-based description of dynamics but do not account for dissipation or external inputs~\citep{goldstein1950classical,greydanus2019hamiltonian}.
We therefore adopt a learned port-Hamiltonian representation to model end-effector dynamics with dissipation and external forcing~\citep{van2004port,zhong2020dissipative,desai2021port,liu2025physically}.
For a system with state $x$ and input $u$, let the Hamiltonian be $\mathcal{H}=T+V$ where $T$ and $V$ denote the system's kinetic and potential energy terms. 
The port-Hamiltonian dynamics therefore takes the form
\begin{equation}
    \dot{x} = \left[J\left(x\right)-D\left(x\right)\right]\nabla\mathcal{H}\left(x\right) + G\left(x\right)u.
\end{equation}
Here, $J$ is the power-preserving symplectic operator, $D$ is the dissipation matrix, and $G$ maps the input into the state dynamics, with constraints $J\left(x\right)=-J\left(x\right)^{\intercal}$ and $D\left(x\right)=D\left(x\right)^{\intercal}\succeq0$.
In the learned surrogate, the required matrix constraints are enforced directly through parameterization rather than penalty terms.
PHRetune restricts the learned representation to end-effector motion rather than explicitly reconstructing the full joint-space dynamics of the manipulator.
This reduced-order representation aligns with the end-effector action interface of the diffusion and VLA policies considered here.
Its Hamiltonian serves as a model-relative storage function for comparing demonstrations with policy-generated motion. We distinguish two models within this framework.
The learned action port model describes the action-to-motion dynamics and defines the demonstration-derived effort and energy envelopes.
The analytical impedance setpoint port model then translates the resulting admissible action scale into fixed controller gains through virtual spring-energy matching.
%%%%%%%%%%%%%%%%%%%%%%%%%%%%%%%%%%%%%%%%%%%%%%%%%%%%%%%%%%%
\subsection{Action Port Model}
PHRetune uses a gray-box neural network to fit a reduced-order port-Hamiltonian action-to-motion surrogate from demonstration transitions.
Let $q$ and $p$ denote the modeled end-effector coordinates and momenta, with state $\mathbf{x}=\left[q, p\right]^{\intercal}$.
We adopt a quadratic Hamiltonian~\citep{mehrmann2023differential},
\begin{equation}
    \mathcal{H}_{\text{a-p}} = \frac{1}{2}\hat{q}^{\intercal}K\hat{q} + \frac{1}{2}\hat{p}^{\intercal}M^{-1}\hat{p},
\end{equation}
where $M$ and $K$ are the surrogate's effective inertia and stiffness matrices.
These parameters define model-relative storage rather than the physical energy parameters of the full manipulator.
The action-driven dynamics take the form
\begin{equation}
    \dot{\mathbf{x}} = \left(J-D\right)\nabla_{\mathbf{x}}\mathcal{H}_{\text{a-p}}\left(\mathbf{x}\right) + G\mathbf{a},
\end{equation}
where $J$ is the symplectic operator, $D$ represents dissipation, and $G=\left[0, G_{u}\right]^{\intercal}$ maps policy actions to momentum dynamics.
The model parameters satisfy
\begin{equation}
    M = M^{\intercal}\succ0, \quad K = K^{\intercal}\succeq0, \quad D = D^{\intercal}\succeq0.
\end{equation}
The power-conjugate output associated with the action port is
\begin{equation}
    y = G^{\intercal}\nabla_{\mathbf{x}}\mathcal{H}_{\text{a-p}}, = G_{u}^{\intercal}\nabla_{p}\mathcal{H}_{\text{a-p}} = G_{u}^{\intercal}M^{-1}p.
\end{equation}
Using the skew-symmetry of $J$, the storage derivative becomes
\begin{equation}
    \dot{\mathcal{H}}_{\text{a-p}}=-\left(\nabla_{\mathbf{x}}\mathcal{H}_{\text{a-p}}\right)^{\intercal}D\nabla_{\mathbf{x}}\mathcal{H}_{\text{a-p}}+\mathbf{a}^{\intercal}y \leq \mathbf{a}^{\intercal}y.
\end{equation}
The continuous-time surrogate is passive with respect to the action–output port, meaning that its stored energy cannot increase by more than the energy supplied through its action input.
When PHRetune compares predicted storage increases against demonstration-derived budgets, those increases therefore reflect the supplied actions rather than energy created by the surrogate itself.
Given demonstration transitions
\begin{equation}
    \mathcal{D}_{k} = \left(q_{k},\dot{q}_{k},\mathbf{a}_{k},q_{k+1},\dot{q}_{k+1}\right),
\end{equation}
we identify the model parameters by predicting the observed motion through the transition map
\begin{equation}
    \hat{\mathbf{x}}_{k+1} = \Phi_{\Delta t}\left(\mathbf{x}_{k}, \mathbf{a}_{k}\right),
\end{equation}
where $\Delta t$ is the sampling interval and hat denotes surrogate predicted quantities.
For an $N$-step demonstration window, the fitting loss is
\begin{equation}
    \mathcal{L} = \sum^{N}_{j=1}\left[\left\lVert\frac{\hat{q}_{k+j}-q_{k+j}}{\Delta t}\right\rVert^{2} + \left\lVert M^{-1}\hat{p}_{k+j}-\dot{q}_{k+j}\right\rVert^{2}\right], 
\end{equation}
The first term penalizes position prediction error normalized by the sampling interval, while the second compares the model’s velocity prediction with the recorded velocity. 
Cholesky-factor parameterizations enforce the definiteness constraints throughout identification.
We train PHRetune using 10\% of demonstrations used to train the base policy.

The fitted surrogate enables comparison between demonstration motion with the frozen policy’s predictions.
We perform this assessment in shadow mode by applying the policy to recorded demonstration observations and propagating its predicted action chunks through the surrogate from the corresponding recorded initial states.
Let $\tau$ denote the number of actions in a chunk, the predicted states are obtained from
\begin{equation}
    \hat{\mathbf{x}}_{j+1} = \Phi_{\Delta t}\left(\hat{\mathbf{x}}_{j}, \hat{\mathbf{a}}_{j}\right), \quad j=0,\ldots,\tau-1.
\end{equation}
Notably, no policy-generated action is executed on the robot or simulator during this assessment.
We first define a demonstration-derived effort budget using the empirical 90th percentile of the mapped action magnitudes,
\begin{equation}
    u_{\max} = Q_{0.9}\left\{\lVert G_{u}\mathbf{a}_{k}\rVert_{2}\right\},
\end{equation}
where the percentile is computed over recorded demonstration actions. 
For a positive effort budget, the normalized effort score of a policy-generated chunk is
\begin{equation}
    U^{\mathrm{eff}}_{\Phi}\left(\tau\right) = \frac{\max_{0<k<\tau}\lVert G_{u}\hat{\mathbf{a}}_{k}\rVert_{2}}{u_{\max}}.
\end{equation}
This score measures the largest model-relative commanded effort within the chunk relative to the demonstration budget, rather than comparing raw action magnitudes.
Similarly, we define the energy budget from the largest storage increase within each demonstration window of the same horizon,
\begin{equation}
    E_{\mathrm{max}} = Q_{0.9}\left\{\max_{0<k<\tau}\left[\mathcal{H}_{\text{a-p}}\left(\mathbf{x}_{k}\right) - \mathcal{H}_{\text{a-p}}\left(\mathbf{x}_{o}\right)\right]\right\},
\end{equation}
where the percentile is taken across demonstration windows.
The corresponding normalized energy score is
\begin{equation}
    U^{\mathrm{en}}_{\Phi}\left(\tau\right) = \frac{\max_{0<k<\tau}\left[\mathcal{H}\left(\hat{\mathbf{x}}_{k}\right) - \mathcal{H}\left(\hat{\mathbf{x}}_{o}\right)\right]}{E_{\mathrm{max}}}.
\end{equation}
Using linear effort scaling and a quadratic approximation for the energy score, we impose the algebraic envelope constraints
\begin{equation}
    sU^{\mathrm{eff}}_{\Phi}\left(\tau\right) \leq 1, \quad s^{2}U^{\mathrm{en}}_{\Phi} \leq 1.
\end{equation}
The largest scale satisfying these constraints without amplifying the nominal actions is
\begin{equation}
    s^{*} = \mathrm{min}\left\{1.0,\nicefrac{1}{U^{\mathrm{eff}}_{\Phi}\left(\tau\right)}, \nicefrac{1}{\sqrt{U^{\mathrm{en}}_{\Phi}\left(\tau\right)}}\right\}.\label{eqn:phretune_clipping}
\end{equation}
For fixed-gain deployment, we apply this rule to task-level percentile summaries of the effort and energy scores, and clip the resulting action scales before conversion to controller gains.
The upper bound of one restricts the rule to attenuation; passivity of the continuous-time surrogate is established independently by its structural constraints.
%%%%%%%%%%%%%%%%%%%%%%%%%%%%%%%%%%%%%%%%%%%%%%%%%%%%%%%%%%%
\subsection{Impedance Setpoint Port Model}
The action port model identifies the model-relative energy and effort envelopes from demonstration data alone, but the derived action scale cannot be applied directly as a downstream controller gain.
To convert the identified scale into executable gains, we use a second controller-in-the-loop model, termed the impedance setpoint port model.
For the translational impedance loop, let $q$ denote the end-effector position and $r$ the commanded setpoint. 
The tracking error and reduced linear dynamics are
\begin{equation}
    e=r-q, \quad m\ddot{q} = k_{p}e - k_{d}\dot{q},
\end{equation}
where $m$ is a scalar effective mass and $k_{p}$ and $k_{d}$ are fixed scalar proportional and derivative gains.
The corresponding Hamiltonian is
\begin{equation}
    \mathcal{H}_{\text{is-p}} = \frac{1}{2}m\lVert\dot{q}\rVert^{2} + \frac{1}{2}k_{p}\lVert e\rVert^{2},
\end{equation}
with power balance
\begin{equation}
    \dot{\mathcal{H}}_{\text{is-p}} = -k_{d}\lVert\dot{q}\rVert^{2} + \left(k_{p}e\right)^{\intercal}\dot{r}.
\end{equation}
The spring effort and setpoint velocity therefore form the moving-setpoint effort–flow port through which the commanded equilibrium can supply energy to this reduced model.

We use this representation to relate action attenuation to controller softening. For the OSC-Pose controller~\cite{khatib1987unified} considered for the Franka arms, each translational action constructs a setpoint relative to the current end-effector position,
\begin{equation}
    r_{k} = q_{k} + 0.05\, s\, \mathbf{a}_{k},
\end{equation}
where $s$ is the action scale and 0.05 is the controller’s action-to-displacement factor. 
Let $\beta$ scale the proportional gain. 
At setpoint construction, the associated virtual spring energy is
\begin{equation}
    W_{k}\left(s,\beta\right) = \frac{1}{2}\left(\beta k_{p}\right)\left(0.05\, s\right)^{2}\lVert \mathbf{a}_{k}\rVert^{2}_{2}.
\end{equation}
Action attenuation reduces the commanded displacement while retaining nominal stiffness, whereas gain retuning reduces stiffness while leaving the action unchanged.
Matching their virtual spring energies gives
\begin{equation}
    W_{k}\left(s^{*},1\right) = W_{k}\left(1, \beta^{*}\right),
\end{equation}
and therefore the closed-form map
\begin{equation}
    \beta^{*} = \left(s^{*}\right)^{2}.
\end{equation}
The resulting proportional and derivative gains are
\begin{equation}
    k_{p}' = \beta^{*}k_{p}, \quad k_{d}' = \sqrt{\beta^{*}}k_{d}.
\end{equation}
For a positive gain scale, the derivative-gain adjustment preserves the damping ratio of the reduced model
\begin{equation}
    \zeta' = \frac{k_{d}'}{2\sqrt{k_{p}'m}} = \frac{\sqrt{\beta^{*}}k_{d}}{2\sqrt{\beta^{*}k_{p}m}} = \frac{k_{d}}{2\sqrt{k_{p}m}} = \zeta.
\end{equation}

For comparison, we also derive an alternative mapping that matches virtual spring force rather than virtual spring energy.
The spring-force magnitude at setpoint construction is
\begin{equation}
    F_{k}\left(s,\beta\right) = \left(\beta k_{p}\right)\left(0.05\, s\right)\lVert \mathbf{a}_{k}\rVert_{2}.
\end{equation}
Matching action-side attenuation and gain-side retuning yields
\begin{equation}
    \beta^{\dagger} = s^{*}.
\end{equation}
The corresponding gains follow the same damping-ratio-preserving update. 
For action scales strictly between zero and one, energy matching produces a softer controller than force matching. 
The two mappings are compared in the LIBERO gain-sweep experiments in Section~\ref{subsec:exp_diffusionpolicy}.
%%%%%%%%%%%%%%%%%%%%%%%%%%%%%%%%%%%%%%%%%%%%%%%%%%%%%%%%%%%
%%%%%%%%%%%%%%%%%%%%%%%%%%%%%%%%%%%%%%%%%%%%%%%%%%%%%%%%
\section{Experiments}\label{sec:experiments}
%%%%%%%%%%%%%%%%%%%%%%%%%%%%%%%%%%%%%%%%%%%%%%%%%%%%%%%%
We evaluate PHRetune through a series of experiments designed to test whether a controller gain computed from demonstrations can improve a frozen Diffusion Policy without empirical gain search.
We compare the derived gain against nominal and principled tuning baselines, assess whether it lies on the empirically optimal gain plateau, and examine its consistency across data regimes.
Unless otherwise stated, all gains are computed once before evaluation and applied only to the downstream controller; the policy, action representation, and inference procedure remain unchanged.
%%%%%%%%%%%%%%%%%%%%%%%%%%%%%%%%%%%%%%%%%%%%%%%%%%%%%%%%
\subsection{Simulated Manipulation with Diffusion Policy}
\label{subsec:exp_diffusionpolicy}
We first evaluate PHRetune with Diffusion Policy on the LIBERO-Goal, LIBERO-Object, and LIBERO-Spatial suites.
We consider manipulation policies trained with either 20 or the full 50 demonstrations for a single-task.
\begin{table}[b]
    \caption{\textbf{Simulation tuned gains.} We report tuned $\beta$ scaling for PHRetune, Inv-$\Sigma$, and DiffTune. Robosuite defaults $k_{p}=150$ with $k_{d}$ coupled via a fixed $1$ damping ratio. The equivalent $\beta$ values for Inv-$\Sigma$ and DiffTune are calculated after their respective derivation.}
    \label{tab:gain_libero}
    \centering
    \begin{tabular}{llccc}
        \toprule
        Backbone & Model & Goal & Obj. & Spa.\\
        \midrule
        \multirow{3}{*}{Diffusion Policy} & Inv-$\Sigma$ & $0.229$ & $0.232$ & $0.244$\\
        & DiffTune & $1.123$ & $1.140$ & $1.131$\\
        & PHRetune & $0.699$ & $0.461$ & $0.531$\\
        \bottomrule
    \end{tabular}
\end{table}
Each single-task policy is evaluated on 50 held-out initial states over three seeds, and we report success rate averaged across tasks and seeds.
The nominal Diffusion Policy (DP) uses the default operational-space controller gains of the Franka Panda manipulator implemented in robosuite~\citep{zhu2020robosuite}.
We compare PHRetune with inverse-covariance derivation (Inv-$\Sigma$)~\citep{calinon2010learning}, model-based controller autotuning method DiffTune~\citep{cheng2024difftune} (DiffTune has access to the ground truth model for the simulator experiments), and intent–execution mismatch retraining method MtG~\citep{xu2026mind} (we re-implement the MtG-SM2M Dual-State conditioning variant).
%
% \footnote{The source code of MtG was not publicly released at the time of submission. We re-implemented MtG-SM2M to be best of our ability.}.
%
Both Diffusion Policy and MtG are trained to 1500 epochs with effective batch size 64.
For PHRetune, the diffusion-policy checkpoint, action representation, and inference procedure remain unchanged, and only the downstream impedance gains are modified in robosuite.
The derived $\beta$ scaling factor is reported in Table~\ref{tab:gain_libero}.
All performance metrics are reported across three seeds (72, 172, and 272).
%%%%%%%%%%%%%%%%%%%%%%%%%%%%%%%%%%%%%%%%%%%%%%%%%%%%%%%%
\subsubsection{PHRetuned Policy Performance}
We compared the success rate of PHRetune against both vanilla Diffusion Policy and baseline methods in Table~\ref{tab:libero}, where PHRetune achieves the highest success rate under both training data regimes.
\begin{table}[t]
    \caption{\textbf{Diffusion Policy success rate.} We report the single-task success rate on LIBERO-goal, -object, and -spatial subsets trained on 20 and 50 demos. PHRetuned Diffusion Policy achieves the highest success rate compared to baseline methods. Notably, PHRetuned Diffusion Policy trained on 20 demos approaches the performance of vanilla Diffusion Policy trained on the full 50.}
    \label{tab:libero}
    \centering
    \begin{tabular}{llcccc}
        \toprule
        Demos & Model & Goal & Obj. & Spa. & Total\\
        \midrule
        \multirow{5}{*}{$20$} & Diffusion Policy & $84.6$ & $76.7$ & $74.3$ & $78.5$\\
        & Inv-$\Sigma$ & $77.1$ & $\underline{\smash{77.5}}$ & $\underline{\smash{81.7}}$ & $\underline{\smash{78.8}}$\\
        & DiffTune & $80.9$ & $71.5$ & $67.9$ & $73.4$\\
        & MtG-SM2M & $\underline{\smash{86.1}}$ & $66.7$ & $80.9$ & $77.9$\\
        & PHRetune & $\mathbf{89.0}$ & $\mathbf{86.4}$ & $\mathbf{88.3}$ & $\mathbf{87.9}$\\
        \midrule
        \multirow{5}{*}{$50$} & Diffusion Policy & $\underline{\smash{92.2}}$ & $\underline{\smash{86.1}}$ & $88.7$ & $\underline{\smash{89.0}}$\\
        & Inv-$\Sigma$ & $83.1$ & $86.0$ & $\underline{\smash{91.2}}$ & $86.8$\\
        & DiffTune &$89.6$ & $83.7$ & $84.9$ & $86.1$\\
        & MtG-SM2M & $87.7$ & $82.1$ & $89.1$ & $86.3$\\
        & PHRetune & $\mathbf{93.9}$ & $\mathbf{89.7}$ & $\mathbf{94.3}$ & $\mathbf{92.6}$\\
        \bottomrule
    \end{tabular}
\end{table}
\begin{figure}[b]
    \centering
    \includegraphics[width=\columnwidth]{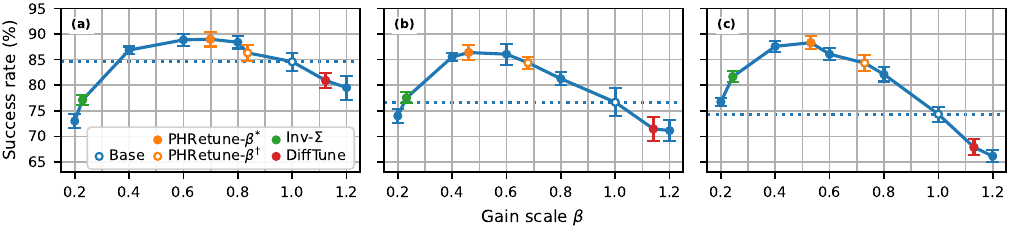}
    \caption{\textbf{LIBERO gain sweep.} We report the success rate over three seeds for the \textbf{(a)} goal, \textbf{(b)} object, and \textbf{(c)} spatial  partition of LIBERO by uniformly sweeping the gain scale. The success rate for PHRetune and baselines are highlighted. MtG is omitted from the curve and $\beta$ values for Inv-$\Sigma$ and DiffTune are calculated post-hoc using derived gain values.}
    \label{fig:PHRetune_GainLadder_LIBERO}
\end{figure}
When trained under 20 demonstrations per task, the success rate of PHRetuned Diffusion Policy approaches the success rate of vanilla Diffusion Policy trained with the full 50 demonstrations.
This illustrates how PHRetune's energy regularization is able to recover lost performance under sparse data regimes.
PHRetune also maintains its utility when given the full 50 demonstrations, improving success rate by 3.6\% compared to the closest baseline.
%%%%%%%%%%%%%%%%%%%%%%%%%%%%%%%%%%%%%%%%%%%%%%%%%%%%%%%%
\subsubsection{Evaluation of PHRetune Gains}
The core claim of PHRetune is that effective controller gains can be derived using inexpensive port-Hamiltonian plant fits, without empirical gain search.
We evaluate the success rate of the frozen 20 demo policies by sweeping the $\beta=\left\{0.2,0.4,0.6,0.8,1.0,1.2\right\}$ in Figure~\ref{fig:PHRetune_GainLadder_LIBERO}.
Across all three LIBERO suites, the energy-based gain scale $\beta^{*}$ achieves the highest observed success rate among the evaluated settings, despite being derived independently of the gain sweep.
Coupled with the low overhead of fitting the port-Hamiltonian plant (approximately three minutes of training on CPU), PHRetune provides a cheap and accurate pathway for improving frozen policies.
Force-based modulation, while improving over vanilla Diffusion Policy, trails energy-based modulation.
Baseline gain selection methods report the post-tuning equivalent $\beta$ values. 
%%%%%%%%%%%%%%%%%%%%%%%%%%%%%%%%%%%%%%%%%%%%%%%%%%%%%%%%
\subsubsection{PHRetuned Energy Consistency}
The method rests on the premise that a policy which over-drives its demonstrations can be returned to their energy manifold, and that a compact port-Hamiltonian plant identified from those demonstrations suffices to determine the required gain.
We verify this against energy measured from the true dynamics rather than from the learned plant used to derive the dose.
The frozen policies are rolled out on all 30 tasks before and after retuning, five episodes each from the canonical initial states.
For each eight-step (0.4 s) action chunk, we measure net task-space controller work in joules by integrating the controller force dotted with the end-effector velocity.
We report controller work relative to each task’s demonstration envelope, with values above 1 indicating excess energy delivery.
The average per-chunk energy falls from 1.47 to 0.93 on Spatial, 1.37 to 1.00 on Goal, and 2.11 to 1.26 on Object after retuning.
Consequently, the retuned policies operate near the energy level of the demonstrations they were trained on.
%%%%%%%%%%%%%%%%%%%%%%%%%%%%%%%%%%%%%%%%%%%%%%%%%%%%%%%%
\subsubsection{Shadow Mode Derivation}
The gains in Table~\ref{tab:gain_libero} are derived in shadow mode.
The frozen policy is applied as a pure function to the demonstration observations, and the effort and energy of its predicted action chunks are scored through the fitted plant against the demonstration budgets, with no hardware or simulator execution.
Shadow mode requires only the demonstrations and one forward pass of the policy over them, which is what makes the derivation practical on hardware.
To verify that the demonstration frames are a sufficient proxy for the policy's own state distribution, we also derive the gains from closed-loop telemetry, rolling the frozen policy out for five log-only episodes per task and scoring its executed chunks the same way.
\begin{table}[t]
    \caption{\textbf{Gain derivation in shadow mode and in closed loop.} Energy-based gain scale $\beta^{*}$ and success rate of the 20-demonstration policies per LIBERO suite. Shadow mode scores the frozen policy's predicted chunks on the demonstration frames alone. Close Loop scores the chunks it executes over simulator rollouts. Both routes recover nearly the same gains and overall success rate.}
    \label{tab:diffusion_gain_derivation}
    \centering
    \begin{tabular}{lccccc}
        \toprule
        \multicolumn{2}{l}{Mode} & Goal & Obj. & Spa. & Total\\
        \midrule
        \multirow{2}{*}{Close Loop} & $\beta^{*}$ & $0.646$ & $0.460$ & $0.546$ & \multirow{2}{*}{$88.0$}\\
        & SR & $89.7$ & $86.3$ & $88.1$ & \\
        \midrule
        \multirow{2}{*}{Shadow Mode} & $\beta^{*}$ & $0.699$ & $0.461$ & $0.531$ & \multirow{2}{*}{$87.9$}\\
        & SR & $89.0$ & $86.4$ & $88.3$ & \\
        \bottomrule
    \end{tabular}
\end{table}
As demonstrated in Table~\ref{tab:diffusion_gain_derivation}, the closed-loop route yields $\beta^{*}$ within $0.05$ of the shadow-mode values.
Furthermore, both derivation modes achieve equal success rate with minor fluctuations between task suites.

A single scalar gain derived from the fitted plant lifts the success rate of every LIBERO suite, recovers most of the gap between 20 and 50 demonstrations, and achieves the highest observed success in an independent gain sweep.
These improvements accompany reduced controller work toward demonstration levels.
Since the derivation needs only the demonstrations and one forward pass of the frozen policy, the same procedure transfers unchanged to hardware execution and other policy backbones.
%%%%%%%%%%%%%%%%%%%%%%%%%%%%%%%%%%%%%%%%%%%%%%%%%%%%%%%%
\subsection{Real-World Manipulation Setup}
\begin{figure}[t]
    \centering
    \includegraphics[width=\columnwidth]{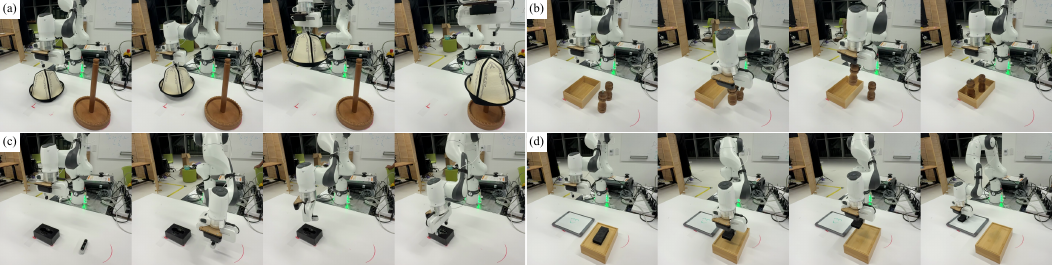}
    \caption{\textbf{Real-world manipulation tasks.} (a) Placing a hat on a stand, (b) transferring salt-and-pepper mills into a tray, (c) placing a camera in its fitted holder, and (d) grasping an eraser and wiping a whiteboard.}
    \label{fig:demo}
\end{figure}
\begin{table}[b]
    \caption{\textbf{Real-world task prompts.} We report the task prompt for all four tasks. Real-world diffusion policy receives no language conditioning; SmolVLA receives language conditioning for single-task policies; and OpenVLA-OFT receives language conditioning for all four tasks.}
    \label{tab:realworld_prompt}
    \centering
    \setlength{\tabcolsep}{2.5pt}
    \begin{tabular}{ll}
        \toprule
        Task & Task Prompt\\
        \midrule
        \texttt{Hat} & \texttt{Place the hat on the stand}\\
        \texttt{S\&P} & \texttt{Put the salt and pepper shakers in the box}\\
        \texttt{Camera} & \texttt{Place the camera into the slot in the box}\\
        \texttt{Eraser} & \texttt{Pick up the eraser and erase the whiteboard}\\
        \bottomrule
    \end{tabular}
\end{table}
We evaluate PHRetune on a common set of four real-world manipulation tasks using Diffusion Policy~\citep{chi2025diffusion}, SmolVLA~\cite{shukor2025smolvla}, and OpenVLA-OFT~\citep{kim2025fine}.
For each backbone, we compare the nominal controller configuration with its PHRetuned counterpart while keeping the policy checkpoint, action representation, and inference procedure unchanged.
The policies are deployed on a Franka FR3 robot arm connected to a Franka Hand manipulator shown in Figure~\ref{fig:demo}, and driven by an OSC-Pose controller implemented in Deoxys~\citep{zhu2022viola} with base $k_{p}$ gain of $350$.
Vision is provided two Orbbec Gemini 335 cameras, one table mounted and one wrist mounted, streaming at 30 FPS.
All real-time inference of baseline policies are run on a server equipped with an Nvidia RTX Pro 6000, Threadripper 7970X, and 256GB of RAM.

We evaluate PHRetune and baseline methods on four real-world manipulation tasks spanning pick-and-place, sequential object manipulation, precision placement, and sustained-contact tool use.
As illustrated in Figure~\ref{fig:demo}, these correspond to placing a hat on its stand (\texttt{hat}), transferring both salt and pepper mills into the tray (\texttt{S\&P}), positioning the realsense camera in its fitted holder (\texttt{camera}), and erasing a whiteboard (\texttt{eraser}), with corresponding task prompts in Table~\ref{tab:realworld_prompt}.
At inference time, success requires the hat to remain on its stand after release, both salt-and-pepper mills to be placed upright inside the tray, the camera to be fully seated flush in its holder, and the target markings to be removed from the whiteboard, respectively.
We manually collect 20 human tele-operated demonstrations recorded at 20Hz for each task for training and evaluate PHRetune and baseline methods over 10 trials that are similarly initialized.
An evaluator blind to the executed method determines success based on the above criteria.
We report the derived gain for Diffusion Policy, SmolVLA, and OpenVLA-OFT in Table~\ref{tab:gain_realworld}.
\begin{table}[t]
    \caption{\textbf{Real-world tuned gains.} We report the derived $\beta^{*}$ for Diffusion Policy, SmolVLA, and OpenVLA-OFT.}
    \label{tab:gain_realworld}
    \centering
    \begin{tabular}{llcccc}
        \toprule
        Backbone & Model & Hat & S\&P & Camera & Eraser\\
        \midrule
        \multirow{3}{*}{Diffusion Policy} & Inv-$\Sigma$ & $0.217$ & $0.178$ & $0.163$ & $0.166$\\
        & DiffTune & $1.109$ & $1.068$ & $1.098$ & $1.104$\\
        & PHRetune & $0.696$ & $0.461$ & $0.707$ & $0.859$\\
        \midrule
        SmolVLA & PHRetune & $0.667$ & $0.460$ & $0.513$& $0.630$\\
        \midrule
        OpenVLA-OFT & PHRetune & $0.731$ & $0.492$ & $0.476$& $0.570$\\
        \bottomrule
    \end{tabular}
\end{table}
%%%%%%%%%%%%%%%%%%%%%%%%%%%%%%%%%%%%%%%%%%%%%%%%%%%%%%%%
\subsection{Real-World Manipulation with Diffusion Policy}
We next evaluate whether the success-rate improvements observed in simulation in Section~\ref{subsec:exp_diffusionpolicy} extend to real-world manipulation with Diffusion Policy.
We apply the same shadow-mode derivation to the real-world demonstrations and frozen policies, computing the controller gains before hardware evaluation without additional rollouts for gain selection.
We compare PHRetune against nominal Diffusion Policy and the same Inv-$\Sigma$, DiffTune, and MtG baselines used in simulation.
\begin{table}[b]
    \caption{\textbf{Real-world manipulation success rate with Diffusion Policy.} We report per-task success rate and their mean across the four tasks. PHRetuned Diffusion Policy outperforms the nominal policy and baseline methods on every task. Tele-Op provides a human teleoperation reference.}
    \label{tab:real_world_diffusion}
    \centering
    \begin{tabular}{lccccc}
        \toprule
        Model & Hat & S\&P & Camera & Eraser & Overall\\
        \midrule
        Tele-Op & 1.00 & 0.87 & 0.87 & 0.91 & 0.91\\
        \midrule
        Diffusion Policy & $\underline{\smash{0.6}}$ & $\underline{\smash{0.4}}$ & $\underline{\smash{0.5}}$ & $\underline{\smash{0.5}}$ & $\underline{\smash{0.50}}$\\
        Inv-$\Sigma$ & $0.2$ & $0.1$ & $0.0$ & $0.0$ & $0.08$\\
        DiffTune & $0.5$ & $0.3$ & $\underline{\smash{0.5}}$ & $0.2$ & $0.38$\\
        MtG-SM2M & $\underline{\smash{0.6}}$ & $\underline{\smash{0.4}}$ & $0.4$ & $0.4$ & $0.45$\\
        PHRetune & $\boldsymbol{0.8}$ & $\boldsymbol{0.6}$ & $\boldsymbol{0.7}$ & $\boldsymbol{0.8}$ & $\boldsymbol{0.73}$\\
        \bottomrule
    \end{tabular}
\end{table}
Both Diffusion Policy and MtG are trained to 1500 epochs with effective batch size 64, and the latest checkpoint is deployed on the Franka FR3 with gains defined in Table~\ref{tab:gain_realworld}.
%%%%%%%%%%%%%%%%%%%%%%%%%%%%%%%%%%%%%%%%%%%%%%%%%%%%%%%%
\subsubsection{Real-World PHRetuned Diffusion Policy Performance}
As illustrated in Table~\ref{tab:real_world_diffusion}, PHRetuned Diffusion Policy outperforms the nominal policy and all baseline methods on each of the four real-world tasks, increasing overall success from $0.50$ to $0.73$.
In contrast, none of the baseline methods improve overall success over nominal Diffusion Policy.
The gain comparison further indicates that controller softening alone does not ensure improved performance.
Both PHRetune and Inv-$\Sigma$ reduce the nominal controller gains, yet Inv-$\Sigma$ decreases success on every task while PHRetune consistently improves it.
These results extend the findings in simulation to real-world manipulation, supporting demonstration-calibrated controller retuning as a means of improving a frozen policy without retraining or additional rollouts for gain selection.
%%%%%%%%%%%%%%%%%%%%%%%%%%%%%%%%%%%%%%%%%%%%%%%%%%%%%%%%
\subsection{Real-World Manipulation with SmolVLA}
To examine whether the improvements extend beyond Diffusion Policy, we next evaluate PHRetune with the vision-language-action model SmolVLA. 
We fine-tune the public 0.45B SmolVLA base model separately for each task with the LeRobot recipe, training the 100M-parameter action expert while the vision-language backbone stays frozen.
\begin{table}[t]
    \caption{\textbf{Real-world manipulation success rate with SmolVLA.} We report per-task success rate across the four tasks. PHRetuned SmolVLA improves success over the nominal policy and DiffTune on every task.}
    \label{tab:real_world_smolvla}
    \centering
    \begin{tabular}{lccccc}
        \toprule
        Model & Hat & S\&P & Camera & Eraser & Overall\\
        \midrule
        SmolVLA & $0.7$ & $0.5$ & $0.4$ & $0.6$ & $0.58$\\
        DiffTune & 0.7 & 0.3 & 0.3 & 0.5 & 0.45\\
        PHRetune & $\boldsymbol{0.9}$ & $\boldsymbol{0.7}$ & $\boldsymbol{0.7}$ & $\boldsymbol{0.9}$ & $\boldsymbol{0.80}$\\
        \bottomrule
    \end{tabular}
\end{table}
\begin{table}[b]
    \caption{\textbf{Real-world motion dynamics with VLAs.} We report end-effector and joint motion metrics, time to grasp, controller gain, and success rate for nominal and PHRetuned SmolVLA, PHRetuned OpenVLA-OFT over ten inference rollouts of the \texttt{hat} task. PHRetune reduces acceleration and jerk toward teleoperation levels with little increase to the time to grasp. Failures do not contribute to the reported metrics. End effector (EE) jerk is measured in m s$^{-3}$, EE accel. in m s$^{-2}$, joint accel. in rad s$^{-2}$, and time to grasp in seconds.}
    \label{tab:real_world_smolvla_dynamics}
    \centering
    \begin{tabular}{lccc}
        \toprule
        Metric & SmolVLA & PHRetune & Tele-Op\\
        \midrule
        EE Jerk RMS & $22.4\pm4.8$ & $11.7\pm4.3$ $\left(\downarrow 47.7\%\right)$ & $9.3\pm1.2$\\
        EE Jerk Peak & $77.2\pm17.4$ & $39.3\pm20.1$ $\left(\downarrow 49.1\%\right)$ & $31.5\pm9.0$\\
        EE Accel. RMS & $0.93\pm0.23$ & $0.52\pm0.17$ $\left(\downarrow 44.1\%\right)$ & $0.43\pm0.07$\\
        EE Accel. Peak & $3.09\pm0.58$ & $1.65\pm0.58$ $\left(\downarrow 46.6\%\right)$ & $1.39\pm0.48$\\
        Joint Accel. RMS & $7.91\pm3.77$ & $3.34\pm1.74$ $\left(\downarrow 57.8\%\right)$ & $2.30\pm0.34$\\
        Time to Grasp & $3.66\pm0.68$ & $3.66\pm0.15$ & $2.99\pm0.48$\\
        $k_{p}$ Gain & 350 & 233 & 350\\
        \midrule
        Success Rate & 0.7 & 0.9 & 1.0\\
        \bottomrule
        \addlinespace[2mm]
        \toprule
        Metric & O.VLA-OFT & PHRetune & Tele-Op\\
        \midrule
        EE Jerk RMS & $19.1\pm5.5$ & $11.6\pm1.0$ $\left(\downarrow 39.3\%\right)$ & $9.3\pm1.2$\\
        EE Jerk Peak & $71.7\pm18.3$ & $42.3\pm2.2$ $\left(\downarrow 41.0\%\right)$ & $31.5\pm9.0$\\
        EE Accel. RMS & $0.97\pm0.24$ & $0.61\pm0.07$ $\left(\downarrow 37.1\%\right)$ & $0.43\pm0.07$\\
        EE Accel. Peak & $3.10\pm0.85$ & $1.78\pm0.25$ $\left(\downarrow 42.6\%\right)$ & $1.39\pm0.48$\\
        Joint Accel. RMS & $6.88\pm2.39$ & $2.26\pm0.46$ $\left(\downarrow 67.2\%\right)$ & $2.30\pm0.34$\\
        Time to Grasp & $3.03\pm0.33$ & $3.12\pm0.39$ & $2.99\pm0.48$\\
        $k_{p}$ Gain & 350 & 256 & 350\\
        \midrule
        Success Rate & 0.8 & 1.0 & 1.0\\
        \bottomrule
    \end{tabular}
\end{table}
The policy receives the two RGB views, an 8-dimensional proprioceptive state and a task prompt, and predicts chunks of 50 actions.
Each model is trained on the same 20 demos used for the diffusion policy and the plant fit, for 50k steps with batch size of $64$ using the recipe's learning rate of $10^{-4}$ under warmup and cosine decay.
Gains in Table~\ref{tab:gain_realworld} are derived from the fine-tuned policy in shadow mode.
%%%%%%%%%%%%%%%%%%%%%%%%%%%%%%%%%%%%%%%%%%%%%%%%%%%%%%%%
\subsubsection{Real-World PHRetuned SmolVLA Performance}
As illustrated in Table~\ref{tab:real_world_smolvla}, PHRetuned SmolVLA improves success on all four real-world tasks, increasing overall success from $0.58$ to $0.80$.
These results extend the findings from Diffusion Policy to SmolVLA, showing that demonstration-calibrated controller retuning also improves the execution of a frozen, language-conditioned policy.
Notably, unlike in LIBERO, DiffTune suffers from degraded performance as it no longer has access to ground truth dynamics to calculate gains from.
%%%%%%%%%%%%%%%%%%%%%%%%%%%%%%%%%%%%%%%%%%%%%%%%%%%%%%%%
\subsubsection{Motion Smoothness of PHRetuned SmolVLA}
We further examine how controller retuning affects the executed motion by comparing end-effector jerk and acceleration, joint acceleration, and time to grasp in Table~\ref{tab:real_world_smolvla_dynamics}.
Reducing the proportional gain from $350$ to $233$ decreases both RMS and peak end-effector jerk by approximately half.
End-effector acceleration also decreases in both measures, while joint acceleration RMS falls by $57.8\%$.
All reported acceleration and jerk measures move closer to the teleoperation reference after retuning.
Notably, these reductions are achieved without a notable increase in the reported time to grasp.
%%%%%%%%%%%%%%%%%%%%%%%%%%%%%%%%%%%%%%%%%%%%%%%%%%%%%%%%
\subsection{Real-World Manipulation with OpenVLA-OFT}
We next evaluate PHRetune with OpenVLA-OFT, extending the SmolVLA experiments to a larger pretrained VLA that predicts action chunks through direct regression rather than flow matching.
\begin{table}[t]
    \caption{\textbf{Real-world manipulation success rate with OpenVLA-OFT.} We report per-task success rate across the four tasks. PHRetuned OpenVLA-OFT improves success over the nominal policy and DiffTune on every task.}
    \label{tab:real_world_openvla}
    \centering
    \begin{tabular}{lccccc}
        \toprule
        Model & Hat & S\&P & Camera & Eraser & Overall\\
        \midrule
        OpenVLA-OFT & $0.8$ & $0.4$ & $0.5$ & $0.7$ & $0.60$\\
        DiffTune & 0.8 & 0.3 & 0.4 & 0.5 & 0.50\\
        PHRetune & $\boldsymbol{1.0}$ & $\boldsymbol{0.6}$ & $\boldsymbol{0.6}$ & $\boldsymbol{0.9}$ & $\boldsymbol{0.78}$\\
        \bottomrule
    \end{tabular}
\end{table}
This evaluation tests whether demonstration-calibrated controller retuning remains effective across differences in model scale and action generation.
We finetune OpenVLA~\citep{kim2024openvla} with OpenVLA-OFT recipe~\citep{kim2025fine} with Rank-32 LoRA~\citep{hu2021lora} adapters on the 7B backbone~\citep{touvron2023llama} with the L1-regression action head and proprioceptive projector trained from scratch.
The policy receives the two RGB views at native resolution, a 7-dimensional proprioceptive state and a task prompt, and predicts chunks of 8 actions.
One finetuned model serves all tasks, conditioned on one fixed prompt per task, trained until loss plateaus at $0.01$ with learning rate $5\times10^{-4}$ and batch size $64$.
%%%%%%%%%%%%%%%%%%%%%%%%%%%%%%%%%%%%%%%%%%%%%%%%%%%%%%%%
\subsubsection{Real-World PHRetuned OpenVLA-OFT Performance}
As illustrated in Table~\ref{tab:real_world_openvla}, PHRetuned OpenVLA-OFT improves success on all four real-world tasks, increasing overall success from 0.60 to 0.78. These results extend the findings from SmolVLA to a larger, regression-based VLA, showing that demonstration-calibrated controller retuning also improves the execution of a single policy shared across all tasks.
%%%%%%%%%%%%%%%%%%%%%%%%%%%%%%%%%%%%%%%%%%%%%%%%%%%%%%%%
\subsubsection{Motion Smoothness of PHRetuned OpenVLA-OFT}
As with the SmolVLA results, Table~\ref{tab:real_world_smolvla_dynamics} shows that PHRetuned OpenVLA-OFT reduces all reported acceleration and jerk measures toward teleoperation levels. 
These reductions are accompanied by a small 3\% increase in time to grasp.
%%%%%%%%%%%%%%%%%%%%%%%%%%%%%%%%%%%%%%%%%%%%%%%%%%%%%%%%

%%%%%%%%%%%%%%%%%%%%%%%%%%%%%%%%%%%%%%%%%%%%%%%%%%%%%%%%
\section{Conclusion}\label{sec:conclusion}
%%%%%%%%%%%%%%%%%%%%%%%%%%%%%%%%%%%%%%%%%%%%%%%%%%%%%%%%
We introduced PHRetune, an offline method that derives controller gains for frozen manipulation policies from their task demonstrations. 
A demonstration-calibrated port-Hamiltonian surrogate assesses policy predictions against effort and energy budgets, while a linear impedance model maps the resulting admissible action scale to fixed stiffness and damping gains.
Gain selection remains blind to evaluation outcomes, requiring neither gain search nor additional robot or simulator rollouts.
The resulting controller retuning preserves the policy, action representation, and inference procedure, introducing no additional runtime computation.

Across LIBERO suites, PHRetune improved Diffusion Policy success, with the independently derived gains achieving the highest observed success rates.
On four real-world manipulation tasks, PHRetuned Diffusion Policy outperformed the nominal policy and all evaluated baselines.
The same procedure improves SmolVLA and OpenVLA-OFT success on all tasks, while reducing acceleration and jerk.
Together, these findings show that task demonstrations can inform not only policy learning but also the controller that executes the learned behavior, providing a practical route to improved manipulation without changing the policy itself.
%%%%%%%%%%%%%%%%%%%%%%%%%%%%%%%%%%%%%%%%%%%%%%%%%%%%%%%%
%%%%%%%%%%%%%%%%%%%%%%%%%%%%%%%%%%%%%%%%%%%%%%%%%%%%%%%%
\bibliographystyle{IEEEtran}
\bibliography{ref}
\end{document}